\documentclass[10pt]{article} 

\usepackage[accepted]{rlj}           

\usepackage{config}
\usepackage{lipsum}
\usepackage{tcolorbox}
\tcbuselibrary{skins} 
\tcbuselibrary{listingsutf8}
\newtcblisting{pseudocodebox}[2]{
enhanced,
    width=#1\linewidth,
    height=#2,
  colback=gray!10,
  colframe=gray!10,
  sharp corners,
boxrule=0pt,
  boxsep=0pt,
  top=0mm,
  bottom=0mm,
  left=1mm,
  right=1mm,
  before skip=1ex,
  after skip=1ex,
  listing only,
  listing options={
    basicstyle=\ttfamily\small,
    breaklines=true,
    columns=fullflexible,
    keepspaces=true,
    aboveskip=0pt, belowskip=0pt,
    escapeinside={(*}{*)},
    texcl=true
  }
}
\newcommand{\boldtt}[1]{\textbf{\texttt{#1}}}

\title{Reinforcement Learning with Physics-Informed Symbolic Program Priors for Zero-Shot Wireless Indoor Navigation}

\setrunningtitle{Physics-Informed Symbolic Priors for Zero-Shot Reinforcement Learning}

\author{Tao Li\textsuperscript{1,$\dagger$}, Haozhe Lei\textsuperscript{1}, Mingsheng Yin\textsuperscript{1}, Yaqi Hu\textsuperscript{1}}

\emails{\{taoli, hl4155, my1778, yh2829\}@nyu.edu}

\affiliations{
$^{1}$\textbf{Department of Electrical and Computer Engineering, New York University}\\
\vspace{-1em} 

$^\dagger$ Corresponding author
}

\begin{document}

\maketitle  

\begin{abstract}
When using reinforcement learning (RL) to tackle physical control tasks, inductive biases that encode physics priors can help improve sample efficiency during training and enhance generalization in testing. However, the current practice of incorporating these helpful physics-informed inductive biases inevitably runs into significant manual labor and domain expertise, making them prohibitive for general users. This work explores a symbolic approach to distill physics-informed inductive biases into RL agents, where the physics priors are expressed in a domain-specific language (DSL) that is human-readable and naturally explainable. Yet, the DSL priors do not translate directly into an implementable policy due to partial and noisy observations and additional physical constraints in navigation tasks. To address this gap, we develop a physics-informed program-guided RL (PiPRL) framework with applications to indoor navigation. PiPRL adopts a hierarchical and modularized neuro-symbolic integration, where a meta symbolic program receives semantically meaningful features from a neural perception module, which form the bases for symbolic programming that encodes physics priors and guides the RL process of a low-level neural controller. Extensive experiments demonstrate that PiPRL consistently outperforms purely symbolic or neural policies and reduces training time by over 26\% with the help of the program-based inductive biases.

\end{abstract}

\section{Introduction} 
Reinforcement learning (RL), while widely explored in a variety of engineering contexts \citep{sim2real-rl,tao22confluence, reddi23rl4sec}, typically suffers from poor sample efficiency in training and limited generalization in testing \citep{mohanty21rl-sample}, especially when facing sophisticated control tasks with high-dimensional sensor inputs and complex system dynamics \citep{lesort18state-rep-control, lauri23pomdp-robot, tao24col, kim-tao25col}. Fortunately, even a small amount of prior knowledge encoded through inductive biases about the task and environment, often seemingly obvious, can dramatically improve learning \citep{baxter00bias, silver19bias}.

While a universally agreed-upon taxonomy of inductive biases in RL has yet to emerge, some common choices are as follows. 1) \textit{Representation bias}, which bears the same spirit of data augmentation, embeds underlying prior knowledge into the training data \citep{han22ramp-meter, zhao22ssd, tao23sce}; 2) \textit{Objective bias} modulates RL processes by shifting the objective function through, for instance, reward shaping \citep{ng99reward-shaping, levine22reward-shaping}, regularization \citep{geist19entropy-mdp, pan-tao24mirror-play}, intrinsic motivation and curiosity \citep{kulkarni16intrinsic, jaques2019social-intrinsic}; 3) \textit{Architectural bias}, which is embedded in the design of the neural networks, leads to a physics-informed architecture \citep{perdikaris19pinn, piccialli22pinn-review} when handling data and environments with physics constraints; and 4) \textit{Algorithmic bias} points to the setup of learning algorithms, including hyperparameters \citep{silver19bias}, initialization \citep{fallah_sgmrl,pan-tao23meta-sg, tao2024meta}, and regularization \citep{mei20softmax,kakade21pg,  pan-tao23delay}.  

Although these inductive biases yield encouraging successes in both theoretical and applied domains, their designs and practical implementations, despite their distinct characteristics, all require intensive manual labor. This problem is even more exacerbated in robotic control tasks, where high-dimensional, noisy sensor readings, sophisticated system dynamics and environments, and physical constraints make it complicated and sometimes painful to introduce inductive biases to learning-based control. One example of such is the wireless indoor navigation task reviewed in \Cref{sec:related}.

\begin{figure}
    \centering
    \includegraphics[width=1\linewidth]{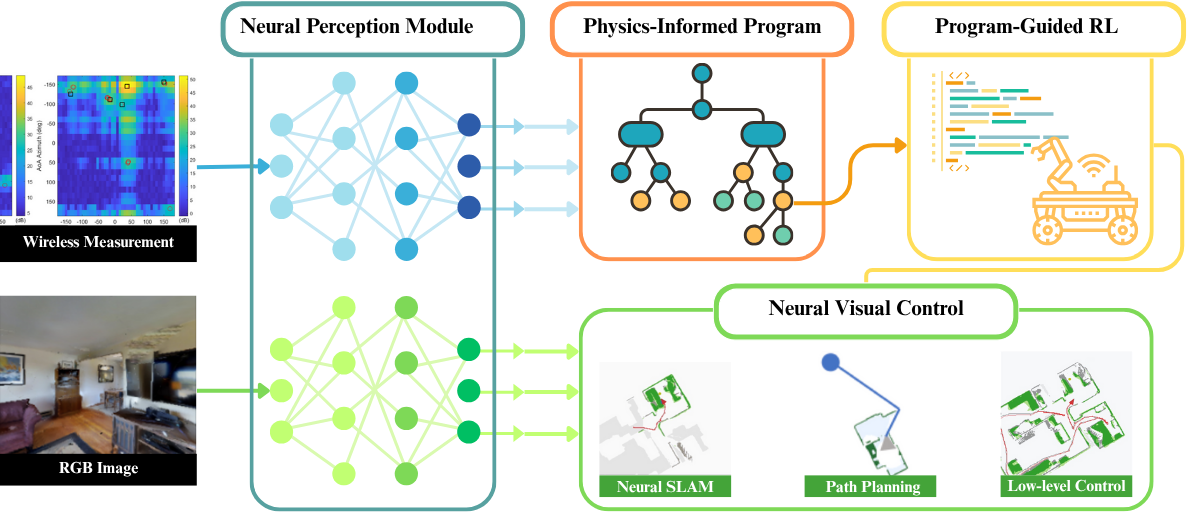}
    \caption{\footnotesize The workflow of the physics-informed program-guided RL (PiPRL). The core of PiPRL is the physics-informed symbolic program, which coordinates the other modules in the wireless indoor navigation tasks.}
    \label{fig:neuro-symbolic}
    \vspace{-0.5cm}
\end{figure}

To facilitate the distillation of inductive biases in RL for real-world robotic tasks, this work proposes to encode physics priors into symbolic programs written in a domain-specific language (DSL) \citep{rodriguez-sanchez23rlang} as the inductive bias, which are human-readable and naturally explainable and can be easily generated from natural languages without handcrafting. Given the close structural alignment between natural language expressions and symbolic programs, our intuition is
\vspace{-0.7\baselineskip}
\begin{center}
    \it when physics priors are easily accessible to humans in natural language,\\ so should it be to agents through symbolic programming languages.
\end{center}
\vspace{-0.7\baselineskip}
Despite its ease of use and inherent explainability, the symbolic program, grounded in abstract physics knowledge, can only provide agents with high-level abstract navigation strategies, such as ``moving along the wireless signal path'', without directly engaging with the low-level sensory signals or motor control needed for real-world robotic operation, for which RL becomes essential.

To get the best of two worlds, we develop a \underline{P}hysics-\underline{i}nformed \underline{P}rogram-guided \underline{R}einforcement \underline{L}earning (PiPRL) framework to achieve zero-shot generalization in wireless indoor navigation tasks (see \Cref{sec:setup}). As shown in \Cref{fig:neuro-symbolic}, our neuro-symbolic RL framework consists of three components: 1) a pretrained \textbf{neural perception module} for processing raw sensor readings, 2) a \textbf{symbolic program module} that expresses physics priors in DSL, which i) directly maps the processed wireless signals to high-level navigation strategies based on physics principles of mmWave propagation and ii) modulate an RL process to search for navigation strategies when the physics priors do not prescribe a executable policy but rather desiderata or constraints that characterize effective policies, and 3) a pretrained low-level RL motion control module translating high-level navigation strategies into control commands.

As an attempt at the \textbf{hierarchical and modularized integration} of neuro policies and symbolic programs in a robotic control task, this work's \textbf{contributions} are as follows. 1) We develop PiPRL to encode physics priors through a symbolic DSL program that serves as an inductive bias for RL processes. We position our work by reviewing the literature in \Cref{app:related}. 2) Besides directly prescribing symbolic policies, PiPRL also employs programs to guide RL policy learning by substituting physics-compliant actions with their unbiased reward estimates. 3)  We empirically demonstrate that PiPRL outperforms purely symbolic and RL policies consistently in terms of sample efficiency and generalizability in Gibson testbeds \citep{xia2018gibson}.

\section{Related Works on Indoor Navigation and Limitations}
\label{sec:related}
 The mobile robot agent aims to utilize the fine-grained temporal and angular resolution of mmWave signal paths to achieve wireless-based positioning and localization \citep{guidi14wireless-slam}, which offers a unique advantage of penetrating beyond line of sight over vision-based methods \citep{habitat, chaplot20neural-slam}. The core challenge of WIN lies in its call for zero-shot generalization, where agents need to complete navigation tasks without fine-tuning when deployed in unseen testing environments.

According to radio frequency propagation, a simple physics-based heuristic of following the mmWave's angle of arrival (AoA) has proven effective and generalizable in structured testbeds \citep{yin2022millimeter}. However, it fails to handle complex environments where mmWave signals propagate along multiple paths through reflections
and diffractions, rendering the observations of signal paths highly noisy and inexact \citep{shahmansoori18wireless-position}. Towards this end, \cite{sutera2020wideband-navi} handcrafted the state representation of wireless sensor measurements (\textit{representation bias}) to indicate the nearest obstacles to the RL agent; \cite{bharadia20deep-loc} developed a hierarchical-decoder network (\textit{architectural bias}) for wireless localization; and most recently, \cite{yin-tao24pirl} and \cite{tao25dt-pirl}  explored an end-to-end deep RL approach with carefully designed reward functions (\textit{objective bias}). While empirically successful, these existing attempts are ad hoc by nature and lead to black-box policies requiring additional efforts for interpretability \citep{tao25dt-pirl}.

\section{Task Setup and Preliminary}
\label{sec:setup}

\textbf{Wireless Indoor Navigation.} Consider a wireless indoor navigation (WIN) task setup as studied in \citep{yin2022millimeter}, where a stationary target is positioned at an unknown location in an indoor environment. The target is equipped with an mmWave transmitter that broadcasts wireless signals at regular intervals. Equipped with an mmWave receiver, an RGB camera, and motion sensors, a mobile robot agent aims to navigate to the target in minimal time. In contrast to the PointGoal task \citep{anderson18evaluate}, WIN does not provide the agent with the target's relative coordinates. 

Like most robotic control tasks \citep{tao22info,lauri23pomdp-robot}, one can formulate WIN as a partially observable Markov decision process (POMDP). The state corresponds to the agent's actual pose $p=(x, y, \varphi)$, where $x, y$ denote the $xy$-coordinate of the agent measured in meters, and $\varphi$ represents the orientation of the agent in degrees (measured counter-clockwise from the $x$-axis). The agent aims to locate and navigate to the target (the wireless transmitter) denoted by $(x^*, y^*)$. Following \citep{chaplot20neural-slam}, the agent employs three navigation actions:  $\mathcal{A}=\{a_F, a_L, a_R\}$, where $a_F=(d,0,0)$ denotes the moving-forward command with a travel distance $d=\SI{25}{cm}$, and $a_L=(0,0,\SI{-10}{\degree})$ and $a_R=(0,0,\SI{10}{\degree})$ denote the turn-left and -right by \SI{10}{\degree} commands. 

Mobile robots are typically equipped with motion sensors that estimate the pose at each time step $t\in \N_{+}$, which we denote by $\bar{p}_t=(\bar{x}_t, \bar{y}_t, \bar{\varphi}_t)$. In addition to the sensor readings, the robot also receives a 3-channel RGB camera image $v_t\in \R^{3\times L_1\times L_2}$ ($L_1$ and $L_2$ are the frame width and height) and mmWave measurement (after signal processing) $w_t\in \R^{N_1\times N_2 \times N_3}$, a three-dimensional tensor whose entries pertain to the spatial-temporal correlation among antenna arrays' angles, received power, and delays \citep{yin2022millimeter}. The tuple $o_t=(\bar{p}_t, v_t, w_t)$ constitutes the partial observation of the agent. Denote by $h_t=\{(o_k, a_k)_{k=1}^t, o_t\}$ the historical observations up to time $t$. The agent aims to find a (stochastic) policy that maps the history to a distribution over actions, $\pi(h_t)\in \Delta(\mathcal{A})$, which generates a sequence of actions $\{a_t\}_{t=1}^H$ finding the target withing the horizon $H$. 

\textbf{Deep Reinforcement Learning.} Using the POMDP language, let $c_t=\|x_t-x^*\|^2+\|y_t-y^*\|^2$ be the Euclidean distance (or any
distance metric, e.g., geodesic distance) between the current pose and the target position, the optimal policy corresponds to the minimizer of the expected cumulative cost, referred to as the value function, $\pi^*\in \argmin J(\pi)\triangleq \E_{\pi}[\sum_{t=1}^T c_t|p_1]$. When facing high-dimensional state inputs in continuous spaces, one can leverage the representation learning power of deep neural networks and consider temporal difference learning and Q-learning with value function approximators \citep{Tsitsiklis97TD, tao19multiRL, mnih2015DQN, fanghui22neural-approx}. 

Alternatively, parameterizing the policy through a deep neural network with parameters $\theta\in \R^n$, one can consider directly optimizing the parameterized policy $\pi_\theta$ through the policy gradient method \citep{sutton_PG, konda99two, mnih16a3c, fujimoto18td3, shutian23erm} and policy gradient-based policy optimization methods \citep{schulman15trpo, schulman17ppo, kakade21pg, pan-tao2025maml-zero-lqr}. This work instantiates the proposed framework using proximal policy optimization (PPO) by \cite{schulman17ppo} to stay consistent with PPO-based RL baselines in the experiments. Note that PiPRL, as a modularized neuro-symbolic RL framework, is fundamentally algorithm-agnostic: the symbolic program component operates independently of the underlying RL optimizer. Any policy-based or value-based RL algorithm can be seamlessly integrated into PiPRL.

\textbf{Domain-Specific Language.}  A domain-specific language (DSL) is a formal language tailored to a specific domain with precisely yet narrowly defined syntax and semantics; for instance, \TeX~by \cite{knuth86tex} is a DSL for typesetting. Compared with general-purpose languages (GPL), such as Python \citep{vanrossum1991python}, DSLs trade general expressivity for ease of use and customization in specific domains, which is exemplified by the Planning Domain Description Languages (PDDL) \citep{pddl}, an early attempt to standardize AI planning languages. This work considers RLang as the DSL, which consists of a set of declarations and element types, with each one corresponding to one or more components of POMDP and associated policy classes, such as options \citep{sutton99semiMDP}; and its formal semantics and grammar are in \citep[Appendix A]{rodriguez-sanchez23rlang}. We briefly review the main RLang element types relevant to our navigation tasks. 

RLang uses \boldtt{Factor} to specify the agent's states (MDP) and partial observations (POMDP). In the WIN task, the partial observation includes pose estimate, vision, and wireless sensor measurements: $o=(\bar{p}, v, w)$, for which RLang defines three factors, as shown in \Cref{fig:rlang}. Closely related to \boldtt{Factor} is \boldtt{Feature}, depicting a function of states, which can be helpful when employing additional information processing, feature extraction, and representation learning to reduce the raw input dimensionality. As will be made clear in \Cref{sec:method}, we use deep learning \citep{chaplot20neural-slam} and low-rank tensor decomposition \citep{blum17tensor-decomp} to extract features from motion and wireless sensor measurements. Besides the two basic elements, two core declarations in RLang are \boldtt{Policy} for executing policies and \boldtt{Effect} for describing rewards and transitions, i.e., subsequent factors. \Cref{fig:rlang} provides an instance of the random navigation policy and the movement effect when executing the move-forward action $a_F$. The ensuing section articulates the symbolic program using the two declarations built on physics priors to instruct the RL agent. 
\begin{figure}
\centering
\begin{minipage}{0.4\linewidth}
    \begin{pseudocodebox}{1}{7.5ex}
(*\textbf{Factor}*) pose := ((*$\bar{x}$*), (*$\bar{y}$*), (*$\bar{\varphi}$*))
(*\textbf{Factor}*) vision := (* $v$ *)
(*\textbf{Factor}*) wireless := (* $w$ *) 
\end{pseudocodebox}
\end{minipage}
\hfill
\begin{minipage}{0.55\linewidth}
    \begin{pseudocodebox}{1}{7.5ex}
(*\textbf{Feature}*) pose estimate := ((*$\hat{x}$*), (*$\hat{y}$*), (*$\hat{\varphi}$*))
(*\textbf{Feature}*) path estimate := (*($g$, $\Omega^{\mathrm{rx}}$, $\Omega^{\mathrm{tx}}$)*)
(*\textbf{Feature}*) link state estimate := (*$\hat{\ell}$*) 
\end{pseudocodebox}
\end{minipage}
\begin{minipage}{0.4\linewidth}
    \begin{pseudocodebox}{1}{10ex}
(*\textbf{Policy}*) random move:
    Execute (*$a_F$*) w/ (*\textbf{P}*)(0.33)
    (*\textbf{or}*) Execute (*$a_L$*) w/ (*\textbf{P}*)(0.33)
    (*\textbf{or}*) Execute (*$a_R$*) w/ (*\textbf{P}*)(0.33)    
\end{pseudocodebox}
\end{minipage}
\hfill
\begin{minipage}{0.55\linewidth}
    \begin{pseudocodebox}{1}{10ex}
(*\textbf{Effect}*) move forward:
    (*\textbf{if} $a_F$*) executed:
        pose = pose + (*$a_F$*)
    (*\textbf{Return}*) pose
\end{pseudocodebox}
\end{minipage}
\vspace{-1em}
\caption{\footnotesize Examples of \boldtt{Factor}, \boldtt{Feature}, \boldtt{Policy}, and \boldtt{Effect} in RLang. \boldtt{Factor} and \boldtt{Feature} describe the agent's observations and related features. \boldtt{Policy} and \boldtt{Effect} represent the navigation policy and its consequence.}
\label{fig:rlang}
\vspace{-0.7cm}
\end{figure}

\section{Physics-Informed Program-Guided Reinforcement Learning}
\label{sec:method}
\textbf{Framework Overview.} A PiPRL agent in WIN first relies on a neural perception module for simultaneous localization and mapping (SLAM), mmWave propagation path estimation. This neural-network module transforms the high-dimensional sensor measurements into the agent’s understanding of the surrounding environment, extracting semantically meaningful features, including signal strength, arrival angles, and obstacle locations. 

These features form the basis for symbolic programming that prescribes high-level navigation strategies aligned with physics priors using \boldtt{Policy} in RLang. However, while some physics knowledge directly leads to executable policies, other knowledge only provides desiderata or necessary conditions that characterize good policies. These high-level principles narrow the policy search space but do not pinpoint an exact policy, for which reinforcement learning remains indispensable. Our physics-informed program (PiP) utilizes \boldtt{Effect} to guide an RL process in searching for optimal policies, resulting in PiP-guided RL. Finally, a vision-based neural network controller breaks down high-level strategies into a sequence of navigation actions to avoid collision using visual information. The following discussion primarily addresses our core contribution on PiP and program-guided RL, while briefly touching on other modules which we adapt from previous works \citep{chaplot20neural-slam, yin2022millimeter, tao25dt-pirl}. Detailed neural network architectures are in \Cref{app:setup}.  

\textbf{Neural Perception. } A critical subtask of indoor navigation is mapping and pose estimation, which creates a map representation of surrounding obstacles and corrects pose readings from motion sensors based on additional visual information. We employ the pretrained neural SLAM model in \citep{chaplot20neural-slam}, which provides robustness to the sensor noise during navigation. This SLAM module internally maintains a spatial map $m_t$ and the agent's pose estimate $\hat{p}_t$ that is different from the raw sensor reading $\bar{p}_t$. The spatial map is represented as  $m_t\in [0,1]^{2\times M\times M}$ is a 2-channel $M\times M$ matrix, where $M\times M$ denotes the map size. Each element in the first channel represents the probability of an obstacle at the corresponding location, while those in the second channel denote the probability of that location being explored. Note that each ``location'' in the spatial map corresponds to a cell of size $\SI{5}{cm}\times \SI{5}{cm}$ in the physical world. The neural SLAM module, denoted by $\bm{F}_{\text{SLAM}}$, is parameterized by a residual network ($\theta_{\text{SLAM}}$) \citep{resnet} and generates new pose estimates and maps auto-regressively using current the  RGB image, the two most recent motion sensor readings, and previous maps and pose estimates: $m_{t}, \hat{p}_t=\bm{F}_{\text{SLAM}}(v_t, \bar{p}_{t-1:t}, \hat{p}_{t-1}, m_{t-1}|\theta_{\text{SLAM}})$.

Wireless measurements also require additional processing to extract hidden information about the signal propagation path. The three-dimensional tensor $w_t\in \R^{N_1\times N_2\times N_3}$, after path estimation through low-rank decomposition \citep{blum17tensor-decomp,yin2022millimeter}, produces tuples $\{g_n, \Omega_n^{\mathrm{rx}}, \Omega_n^{\mathrm{tx}}\}_{n=1}^N$, where where $N$ is the maximum number of detected paths along which signals propagate. For the $n$-th path, $g_n$ denotes its signal-to-noise ratio (SNR),  $\Omega_n^{\mathrm{rx}}$ and $\Omega_n^{\mathrm{tx}}$ denote the angle of arrival (AoA) and departure (AoD), respectively. We denote by $(g_1, \Omega_1^{\mathrm{rx}}, \Omega_1^{\mathrm{tx}})$ the tuple under the strongest signal strength, indicating that the path experiences the least number of reflections.

Another quantity crucial to WIN is the link state, which is categorized into Line-of-Sight (LOS) and Non-Line-of-Sight (NLOS). A location $(x,y)$ (or pose $p$) is said to be of LOS if there is a wireless signal path wherein electromagnetic waves traverse from the transmitter to the receiver without encountering any obstacles. In contrast, NLOS signifies the absence of such a direct visual path. NLOS can further be subdivided into first-order, second-order, third-order, and so forth. $X$-order NLOS ($X\in \N_{+}$) implies that at least one electromagnetic wave in the wireless link undergoes $X$-time reflection or diffraction. Note that the link state, denoted by $\ell_t=1,2,\ldots$, is a wireless terminology instead of the actual state input to be fed into RL models. Instead, the agent learns to estimate the link state, $\hat{\ell}_t$, from the path estimates using a vanilla fully connected network  \citep{yin2022millimeter}. In summary, one can view this neural wireless information processing as a neural network-parameterized function: $\{g_n(t), \Omega_n^{\mathrm{rx}}(t), \Omega_n^{\mathrm{tx}}(t)\}_{n=1}^N, \hat{\ell}_t= \bm{F}_{\text{wireless}}(w_t|\theta_{\text{wls}})$.

\textbf{Physics-Informed Program.} We first articulate the physics priors that are helpful in shaping an effective navigation strategy. \textit{Prior}~\#1: The \textbf{principle of reversibility} states that an electromagnetic wave traversing from the source to the target will follow the same path if its direction is reversed. Such a principle leads to a simple yet effective strategy: following the angle of arrival (AoA) of the strongest path, since the strongest path experiences the least number of reflections and naturally detours around obstacles. We create a symbolic policy, referred to as \texttt{reverse AoA}, to encapsulate this strategy, where \texttt{intermediate} is a waypoint along the AoA (the second entry of \texttt{path estimate}) by an offset distance of $D=\SI{2.5}{m}$ (hyperparameter). The \boldtt{Option}, a declaration in RLang, prescribes a sequence of actions starting from the initial condition \boldtt{init} to the termination \boldtt{until}. The \boldtt{Option} represents a low-level vision-based controller to be introduced later.   
\begin{pseudocodebox}{1}{14.5ex}
(*\textbf{Policy}*): reverse AoA
    (*\textbf{if not}*) pose estimate == goal:
        intermediate[1] := pose estimate[1] + D * cos(path estimate[2])
        intermediate[2] := pose estimate[2] + D * sin(path estimate[2])
        Execute (*\textbf{Option}*): Visual Control: (*\textbf{init}*) := pose estimate (*\textbf{until}*) := intermediate
\end{pseudocodebox}    
The reversibility prior is less effective in higher-order NLOS, for which we consider \textit{Prior}~\#2: the source of an electromagnetic wave acquires the \textbf{maximum signal strength}, which declines along the path. In other words, if one considers the overall SNR at a pose $g(p)=\sum_{n=1}^N g_n(p)$, the closer the agent is to the transmitter, the higher its received SNR is. In contrast to \textit{Prior}~\#1, this prior knowledge does not directly prescribe a navigation strategy, since the neural perception does not provide an SNR ascent direction, unlike AoA. It rather lays down a desideratum that the optimal strategy should meet. Similarly, \textit{Prior}~\#3, rooted in the \textbf{link state monotonicity}, states that a necessary condition for a navigation path to be optimal is that the link state decreases monotonically. 

\textbf{Program-Guided RL.} Since \textit{Prior}~\#2 and \#3 do not induce a precise symbolic policy, we turn to PPO, which is referred to as the \boldtt{Neural Policy}, to search for an RL policy that complies with the priors. To be consistent with the symbolic policy \texttt{reverse AoA}, the output of \boldtt{Neural Policy} is also an intermediate waypoint (equivalently, the offset angle $\hat{\Omega}_t$ w.r.t. $x$-axis), and the input is the processed wireless data $\bm{F}_{\text{wireless}}(w_t)$.  In practical implementations, we consider a discretization of the angle range $\hat{\Omega}_t\in \Pi=\{-170, \ldots, 0, \ldots, 170, 180\}$. Substitute into the distance function $c_t$ the coordinate of the waypoint determined by $\hat{\Omega}_t$, and we obtain the reward.      

The central question in guiding the PPO process is how to program those desiderata using RLang. At first glance, it seems that this question bears a similar spirit of guided policy search \citep{levine14guided-learning, tao23cola} and exploration \citep{kang18ppo-lfd, siji23hybrid-ps}, where physics priors provide guidance. However, it is not straightforward to translate such priors into concrete policy, which is a prerequisite for the importance sampling (IS) technique commonly employed in guided search. We consider directly replacing the sampled action from the PPO network with a physics prior-compliant one (if available) and replacing the actual reward with the IS-weighted one \citep{paine20demo,libardi21ppo-lfd-single}. The reweighted rewards only apply to the action loss without changing the target value in the value loss as in \citep{espeholt18action-learner, libardi21ppo-lfd-single}.

Denote by $\pi_\phi$ the neural network policy, and the action selection distribution is given by $\pi_{\phi}(\cdot|\bm{F}_{\text{wireless}}(w_t))$. We now define the set of prior-compliant actions through RLang element \boldtt{ActionRestriction} and the IS-weighted reward through \boldtt{Effect}. Given the current and last pose estimates $\hat{p}_{t-1:t}$ and the associated SNR $g_t=g(\hat{p}_t)$, the angle of the vector (\texttt{movement angle}) connecting two poses is given by $\nu_t=\arctan( \hat{y}_t-\hat{y}_{t-1}/\hat{x}_t-\hat{x}_{t-1})$ measured in degrees with possible $\pm \SI{180}{\degree}$, depending on which quadrant the vector is within. The compliant actions are those along the SNR ascent direction with deviation no more than $\SI{10}{\degree}$ ( discretization resolution). 
\begin{pseudocodebox}{1}{29.5ex}
(*\textbf{ActionRestriction}*) SNR prior:
    (*\textbf{if}*) SNR > Last SNR:
        (*\textbf{Return}*) angle (*\textbf{in} $\Pi$ \textbf{and} movement angle - \SI{10}{\degree} <= angle <= movement angle + \SI{10}{\degree}*) 
    (*\textbf{if}*) SNR <= Last SNR:
        (*\textbf{Return}*) angle (*\textbf{in} $\Pi$ \textbf{and} movement angle - \SI{190}{\degree} <= angle <= movement angle - \SI{170}{\degree}*)
(*\textbf{Effect}*) Cost Correction:
    (*\textbf{Return}*) cost = cost * \# SNR prior *  (*\textbf{P}(\textbf{Neural Policy})*)(angle)
(*\textbf{Effect}*) Link State Prior:
    (*\textbf{if}*) Line State Estimate > Last Link State Estimate:
        (*{terminate} \textbf{Neural Policy} \textbf{and} reset*) 
\end{pseudocodebox}

Once the action set is settled, whenever the action $\hat{\Omega}_t$ from \boldtt{Neural Policy} is not compliant, the PiPRL agent randomly picks one $\hat{\Omega}_t'$ from the restricted set and incurs a cost $c_t$ that is biased with respect to the original action. To obtain an unbiased estimate, one can employ the importance sampling technique often seen in off-policy evaluation \citep{xie19is-ope,tao20causal} and online learning with bandit feedback \citep{csaba_bandit, tao_blackwell}. The probability of choosing the compliant action is $1/\#\text{compliant}$ under the uniform distribution, while its counterpart under \boldtt{Neural Policy} is given by $\pi_{\phi}(\hat{\Omega}_t')$. The quotient ${\pi_{\phi}(\hat{\Omega}_t')}/{(1/\#\text{compliant})}$ accounts for the correction in the change of probability measure. Hence, the corrected cost is $\hat{c}_t=\#\text{compliant}\cdot \pi_{\phi}(\hat{\Omega}_t')\cdot c_t$,as shown in the \boldtt{Effect} above. Finally, to enforce the \textit{Prior}~\#3 on link state monotonicity, we simply define an \boldtt{Effect} that terminates the PPO training if the link state estimate increases and restarts a new episode. We summarize the entire PiPRL workflow in the following, where a meta symbolic program instructs other policy programs and guidance programs. 
\begin{pseudocodebox}{1}{14.5ex}
(*\textbf{Policy}*) meta-program:
    (*\textbf{if}*) link state estimate <= 2:
            Execute (*\textbf{Policy}*): reverse AoA
    (*\textbf{else}*):
        Execute (*\textbf{Neural Policy}*): PPO (*\textbf{and}*) (*\textbf{ActionRestriction}*): SNR prior  (*\textbf{and}*) 
        (*\textbf{Effect}*): Cost Correction, Link State Prior
\end{pseudocodebox}

\textbf{Visual Controller.} Once the intermediate waypoint is determined either by the symolic or neural policy, a path planner denoted by $\bm{F}_{\text{planner}}$, based on the Fast Marching method \citep{sethian96fmm}, computes the shortest path from the current location to the waypoint using the spatial map $m_t$ and the pose estimate $\hat{p}_t$ from the SLAM module. The unexplored area is considered a
free space for planning. The output of the planner is a short-term goal $p_t^S=\bm{F}_{\text{planner}}(p^L_t, m_t, \hat{p}_t)$, which is the farthest point on the path within the map. Then, the visual controller takes in the path-planning output and the camera images, producing navigation actions $a_t=\pi_{\text{ctrl}}(v_t,p_t^S|\theta_{\text{ctrl}})$ for collision avoidance. The visual controller, parameterized by a recurrent neural network consisting of ResNet18 \citep{resnet}, is a task-invariant neural policy pretrained through behavioral cloning \citep{chaplot20neural-slam}.  
\section{Experiments}
This section evaluates the proposed PiPRL for WIN tasks, aiming to answer the following questions. 1) \textbf{Sample Efficiency}: Does PiPRL take fewer training data than the non-physics-based baselines? 2) \textbf{Zero-shot Generalization}: Can the PiPRL agent navigate in unseen environments without fine-tuning? Due to the page limit, we move the ablation study on the SNR and link state priors to \Cref{app:add-exp}. We follow the same setup in \citep{tao25dt-pirl} and briefly touch upon some key aspects. We remind the reader that $\bm{F}_{\text{SLAM}}$, $\bm{F}_{\text{wireless}}$ and $\bm{F}_{\text{planner}}$ are all pre-trained models and the program-guided PPO is our main focus, whose hyperparameters are listed in \Cref{app:setup}. 

\textbf{Experiment Setup.} The experiment includes 21 different indoor maps (the first 15, A--O, for training and the remaining 6, P--U, for testing) from the Gibson dataset \citep{xia2018gibson} labeled using the first 21 characters in the Latin alphabet (A, B, $\ldots$, U). Table \ref{tab:label_map} in \Cref{app:setup} presents the complete list. For each map, we pick ten target locations that are classified into three categories. The first three targets (1-3) are of LOS (i.e., the agent's starting position is within the LOS area), the next three (4-6) belong to $1$-NLOS, and the remaining four (7-10) correspond to $2^+$-NLOS scenarios.  

\textbf{Baselines.} We consider three baseline navigation algorithms: 1) non-physics-based end-to-end RL (NPRL): the RL policy is purely the \boldtt{Neural Policy} without priors. 2) Wireless-assisted navigation (WAN): This non-RL-based method, put forth in \citep{yin2022millimeter}, only utilizes the symbolic policy \boldtt{reverse AoA} in LOS and 1-NLOS, while switching to random exploration in $2^{+}$-NLOS using the neural SLAM as suggested by \citep{chaplot20neural-slam}. 3) Vision-augmented SLAM (V-SLAM): This policy leverages an object-detection algorithm \citep{bochkovskiy2020yolov4} in computer vision: once the target transmitter is within the view, the agent can localize and approach the target. It reduces to neural SLAM otherwise.  The first two are primary baselines since our PiPRL is a hybrid of neural and symbolic policies. Additionally, to highlight the necessity of leveraging wireless signals in indoor navigation, we consider the third baseline where V-SLAM only takes in RGB images without wireless inputs.

\textbf{Training.} The first 15 maps (A-O) with associated 10 task positions are utilized to learn a PiPRL policy in \textit{sequential order}. The training process follows a specific sequence, starting with task A and progressing to A10, followed by training under tasks B1 to B10. Each task consists of 1000 training episodes. One training instance terminates if the agent completes the task more than 6 episodes out of 10. This procedure is repeated until the agent has been exposed to all 15 maps with all target positions. In contrast, NPRL follows \textit{rotation} training to alleviate catastrophic forgetting: after finishing the training on the current task, we randomly select a few previous tasks \citep{tao25dt-pirl} to re-train the model before moving to the next task to refresh NPRL's ``memory''. 

\textbf{Sample Efficiency.}  We first evaluate the sample efficiency of the PiPRL training process by comparing the number of training episodes and GPU hours of PiPRL in LOS, 1-NLOS, and $2^+$-NLOS with those of NPRL. As one can see from \Cref{tab:gpu_trunc} (full table in \Cref{app:add-exp}), the encoding physics priors via symbolic programs does reduce training samples. The \#GPU hours are reduced by 26\%, which is even more so as training progresses, when the neural policy starts to acquire the physics prior guided by the program. 
\begin{table}[!ht]
\vspace{-1ex}
\centering
\caption{\footnotesize GPU hours and Episodes (Eps) of PiPRL and NPRL for Maps A-D (More in \Cref{tab:final_gpu_episodes}). Experiments are conducted on a Linux GPU workstation with an AMD Threadripper 3990X (64 Cores, 2.90 GHz) and an NVidia RTX 8000.}
\resizebox{\linewidth}{!}{
\begin{tabular}{llccccc}
\toprule
\multirow{2}{*}{Label} & \multirow{2}{*}{Map Name} & \multicolumn{2}{c}{PiPRL} & \multicolumn{2}{c}{NPRL} & \multirow{2}{*}{\shortstack{Reduction \\ (GPU Hours\%)}} \\
\cmidrule(r){3-4} \cmidrule(r){5-6}
      &          & GPU Hours & EPs (Average) & GPU Hours & EPs (Average) & \\
\midrule
A & Bowlus      & 20.35 & 798 & 27.70 & 1000 & 26.56 \\
B & Arkansaw    & 17.81 & 579 & 25.32 & 1000 & 29.67 \\
C & Andrian     & 15.21 & 440 & 24.90 &  996 & 38.93 \\
D & Anaheim     & 13.34 & 384 & 25.11 & 1000 & 46.86 \\
\bottomrule
\end{tabular}
}
\label{tab:gpu_trunc}
\vspace{-3ex}
\end{table}

\textbf{Generalization.} We first highlight that our testing environments (new maps with different target positions) are structurally different
from training cases. Different room topologies and wireless source locations create drastically different wireless fields unseen in the training phase, as the reflection and diffraction patterns are distinct across each setup. \Cref{tab:npl-trunc} summarizes part of the testing results (the first three testing maps; the full table is in \Cref{app:add-exp}). We report the average \textit{normalized path length} (NPL) of 20 repeated tests, which is defined as the quotient of the actual path length (the
number of navigation actions) over the shortest path length
of the testing task (the minimal number of actions). The
closer NPL is to 1, the more efficient the navigation is. One can see from \Cref{tab:npl-trunc} that PiPRL consistently outperforms others. 
\begin{table*}[!ht]
\centering
\caption{\footnotesize A comparison of NPLs under 3 testing maps (More in \Cref{tab:npl}) PIRL achieves impressively efficient navigation in the challenging scenario $2^+$-NLOS, compared with baselines. }
\label{tab:npl-trunc}
\fontsize{6.6pt}{8pt}\selectfont
\begin{tabular}{|l|ccc|ccc|ccc|}
\hline
           & \multicolumn{3}{c|}{Map P}                                                                                                                                                                                                            & \multicolumn{3}{c|}{Map Q}                                                                                                                                                                                                            & \multicolumn{3}{c|}{Map R}                                                                                                                                                                                                            \\ \hline
           & \multicolumn{1}{c|}{LOS}                                                        & \multicolumn{1}{c|}{1-NLOS} & \multicolumn{1}{c|}{$2^+$-NLOS} & \multicolumn{1}{c|}{LOS}                                                        & \multicolumn{1}{c|}{1-NLOS} & \multicolumn{1}{c|}{$2^+$-NLOS} & \multicolumn{1}{c|}{LOS}                                                        &\multicolumn{1}{c|}{1-NLOS} & \multicolumn{1}{c|}{$2^+$-NLOS} \\ \hline
PiPRL       & \multicolumn{1}{c|}{\begin{tabular}[c]{@{}c@{}}1.01 \\ $\pm$ 0.01\end{tabular}} & \multicolumn{1}{c|}{\begin{tabular}[c]{@{}c@{}}1.25 \\ $\pm$ 0.02\end{tabular}}    & \begin{tabular}[c]{@{}c@{}}2.13 \\ $\pm$ 0.04\end{tabular}     & \multicolumn{1}{c|}{\begin{tabular}[c]{@{}c@{}}1.01 \\ $\pm$ 0.01\end{tabular}} & \multicolumn{1}{c|}{\begin{tabular}[c]{@{}c@{}}1.37 \\ $\pm$ 0.02\end{tabular}}    & \begin{tabular}[c]{@{}c@{}}2.43 \\ $\pm$ 0.05\end{tabular}     & \multicolumn{1}{c|}{\begin{tabular}[c]{@{}c@{}}1.01 \\ $\pm$ 0.00\end{tabular}} & \multicolumn{1}{c|}{\begin{tabular}[c]{@{}c@{}}1.13 \\ $\pm$ 0.03\end{tabular}}    & \begin{tabular}[c]{@{}c@{}}2.65\\ $\pm$ 0.06\end{tabular}      \\ \hline
NPRL        & \multicolumn{1}{c|}{\begin{tabular}[c]{@{}c@{}}2.03 \\ $\pm$ 1.02\end{tabular}} & \multicolumn{1}{c|}{\begin{tabular}[c]{@{}c@{}}2.72 \\ $\pm$ 0.55\end{tabular}}    & \begin{tabular}[c]{@{}c@{}}4.96 \\ $\pm$ 1.37\end{tabular}     & \multicolumn{1}{c|}{\begin{tabular}[c]{@{}c@{}}2.12 \\ $\pm$ 1.00\end{tabular}} & \multicolumn{1}{c|}{\begin{tabular}[c]{@{}c@{}}3.08 \\ $\pm$ 0.68\end{tabular}}    & \begin{tabular}[c]{@{}c@{}}5.00 \\ $\pm$ 1.41\end{tabular}     & \multicolumn{1}{c|}{\begin{tabular}[c]{@{}c@{}}2.28 \\ $\pm$ 1.03\end{tabular}} & \multicolumn{1}{c|}{\begin{tabular}[c]{@{}c@{}}2.49 \\ $\pm$ 0.81\end{tabular}}    & \begin{tabular}[c]{@{}c@{}}4.99 \\ $\pm$ 1.20\end{tabular}     \\ \hline
V-SLAM    & \multicolumn{1}{c|}{\begin{tabular}[c]{@{}c@{}}1.05 \\ $\pm$ 0.02\end{tabular}} & \multicolumn{1}{c|}{\begin{tabular}[c]{@{}c@{}}1.82 \\ $\pm$ 0.51\end{tabular}}    & \begin{tabular}[c]{@{}c@{}}4.58 \\ $\pm$ 1.12\end{tabular}     & \multicolumn{1}{c|}{\begin{tabular}[c]{@{}c@{}}1.11\\ $\pm$ 0.03\end{tabular}}  & \multicolumn{1}{c|}{\begin{tabular}[c]{@{}c@{}}2.89 \\ $\pm$ 0.73\end{tabular}}    & \begin{tabular}[c]{@{}c@{}}4.89\\ $\pm$ 1.00\end{tabular}      & \multicolumn{1}{c|}{\begin{tabular}[c]{@{}c@{}}1.09 \\ $\pm$ 0.03\end{tabular}} & \multicolumn{1}{c|}{\begin{tabular}[c]{@{}c@{}}1.68\\ $\pm$ 0.6\end{tabular}}      & \begin{tabular}[c]{@{}c@{}}4.68 \\ $\pm$ 1.01\end{tabular}     \\ \hline
WAN & \multicolumn{1}{c|}{\begin{tabular}[c]{@{}c@{}}1.02 \\ $\pm$ 0.00\end{tabular}} & \multicolumn{1}{c|}{\begin{tabular}[c]{@{}c@{}}1.32 \\ $\pm$ 0.02\end{tabular}}    & \begin{tabular}[c]{@{}c@{}}3.88 \\ $\pm$ 0.82\end{tabular}     & \multicolumn{1}{c|}{\begin{tabular}[c]{@{}c@{}}1.01 \\ $\pm$ 0.01\end{tabular}} & \multicolumn{1}{c|}{\begin{tabular}[c]{@{}c@{}}1.63 \\ $\pm$ 0.05\end{tabular}}    & \begin{tabular}[c]{@{}c@{}}3.71 \\ $\pm$ 0.71\end{tabular}     & \multicolumn{1}{c|}{\begin{tabular}[c]{@{}c@{}}1.01 \\ $\pm$ 0.00\end{tabular}} & \multicolumn{1}{c|}{\begin{tabular}[c]{@{}c@{}}1.23 \\ $\pm$ 0.02\end{tabular}}    & \begin{tabular}[c]{@{}c@{}}3.97 \\ $\pm$ 0.83
\end{tabular}   \\ \hline
\end{tabular}
\vspace{-0.5cm}
\end{table*}
\section{Conclusion}
We introduced PiPRL, a neuro-symbolic reinforcement learning framework that integrates physics-informed symbolic programs with neural network-based perception and control to enable zero-shot generalization in wireless indoor navigation tasks. At the core of PiPRL is a domain-specific language (RLang) for encoding symbolic programs that express physics priors--ranging from directly executable policies to high-level desiderata that characterize effective navigation strategies. These symbolic programs serve as structured, interpretable inductive biases that guide the reinforcement learning process when explicit policy specification is infeasible. 
\newpage
\appendix

\section{Related Works on Neuro-Symbolic Learning}
\label{app:related}
We have reviewed recent advancements in physics-informed inductive biases in RL and machine learning in general. Their specific applications in indoor navigation have also been discussed in the main text. This section is devoted to the recent developments on program-guided agents and neuro-symbolic learning.

\textbf{Program-Guided Agent.} There has been a recent surge of interest in learning from languages and instructions \citep{luketina19lang-rl}. Many of them attempt to learn mappings for the semantic meaning of natural language to grounded information for agents \citep{kaplan2017beating-21f, bahdanau2018learning-199}, which can also be viewed as inductive biases. Some other endeavors include natural-language-based task specification \citep{tellex11, fried18} and policy conditioning \citep{zhang20policy-cond}.

However, natural language statements can often be ambiguous to humans, which motivates the usage of formal languages, including logic programming \citep{bastani19spec-lang, jiang19neural-logic} and domain-specific languages (DSLs) \citep{pddl, rodriguez-sanchez23rlang}. Our work leverages the DSL program. Yet, unlike those prior works \citep{pddl, rodriguez-sanchez23rlang, joseph-lim20program-agent}, our program does not focus on task-specification, grounding information, but rather defining a symbolic policy coherent with physics priors.  

\textbf{Neuro-Symbolic RL} The incorporation of DSL programs also distinguishes our work from contemporary efforts on neuro-symbolic RL. One of the common approaches in combining neuro learning and symbolic planning relies on integrating differentiable logic programming with deep RL policies \citep{jiang19neural-logic, gre19, kersting25blend}. The advantage of this approach is that gradient-based optimization techniques apply to both symbolic planning and RL \citep{gre19,delfosse23}. However, it is less user-friendly than DSL programs in terms of readability and accessibility.

The modularized and hierarchical neuro-symbolic integration has also been explored in the literature \citep{joseph-lim20program-agent, Kokel2021, kuo20}. One of the shared characteristics is that a symbolic program takes the role of a high-level planner, while RL fulfills the commands from symbolic planners. Our PiPRL, even though still hierarchical, also employs RL, guided by symbolic programs, to search for high-level strategies, which is motivated by the complexity of real-world applications where no single symbolic policy is sufficient.

\section{Experiment Setup}
\label{app:setup}
This section provides additional setup details. To begin with, \Cref{tab:label_map} presents our map labeling. In addition, we also summarize the neural network architectures employed in this work.
\begin{itemize}
    \item \textbf{PPO Policy} comprises a recurrent neural network architecture, which includes a linear sequential wireless encoder network with two layers, followed by fully connected layers and a Gated Recurrent Unit (GRU) layer \cite{GRU}. Additionally, there are two distinct layers at the end, referred to as the actor output layer and the critic output layer. We follow the standard PPO implementation as in \citep{schulman17ppo} and summarize its hyperparameters in \Cref{tab:ppo}.
    \item \textbf{Visual Control Policy} is constructed using a recurrent neural network architecture. It incorporates a pre-trained ResNet18 \citep{resnet} as the visual encoder, which is followed by fully connected layers and a GRU layer.
    \item \textbf{Neural SLAM} consists of Resnet18 convolutional layers followed by two fully-connected layers, then followed by 3 deconvolutional layers.
    \item \textbf{Link State Classifier} is simply a 2-layer fully-connected network with ReLU activation \citep{hinton10relu}.
\end{itemize}
\begin{table}[!ht]
\centering
\caption{The label-map correspondence.}
\begin{tabular}{llll|ll}
\toprule
Label & Map Name & Label & Map Name & Label & Map Name \\
\midrule
A & Bowlus & I & Capistrano & P & Woonsocket   \\
B & Arkansaw & J & Delton & Q & Dryville  \\
C & Andrian & K & Bolton & R & Dunmor  \\
D & Anaheim & L & Goffs & S & Hambleton \\
E & Andover & M & Hainesburg & T & Colebrook\\
F & Annawan & N & Kerrtown & U & Hometown\\
G & Azusa & O & Micanopy & & \\
H & Ballou & & & &\\
\bottomrule\label{tab:label_map}
\end{tabular}
\end{table}

\begin{table}[!ht]
\caption{PPO Hyperparameters.}
    \centering
    \begin{tabular}{ll}
    \toprule
    \textbf{Hyperparameters}     &  \textbf{Value}\\
    \midrule
    discounting factor     &  0.99\\
    clipping parameter  & 0.2 \\
    weight for value loss  & 0.5\\
    weight for entropy bonus & 0.01\\
    learning Rate  & $3 \times 10^{-4}$ \\
    Batch Size & 64\\
    exploration rate (e.g., epsilon-greedy) & 0.1 \\
    \bottomrule
    \end{tabular}
    \label{tab:ppo}
\end{table}

\section{Additional Experiments}
\label{app:add-exp}
This section supplements additional experimental results. The comprehensive examinations of sample efficiency and generalization are in \Cref{tab:final_gpu_episodes} and \Cref{tab:npl}, respectively. Additionally, we want to answer the question: \textit{To what extent do the symbolic priors help?}

\subsection{Ablation Study}
 We simply erase the corresponding program when conducting the ablation. For example, when studying SNR prior, we remove \boldtt{ActionRestriction} and \boldtt{Effect} on cost correction.  As one can see from \Cref{tab:ablation}, the answer to the question is affirmative, as the SNR ablation returns significantly higher NPLs in $2^+$-NLOS. Similarly,  the third row in \Cref{tab:ablation} indicates that without the link state prior, the agent frequently revisits the high-order NLOS areas in testing, which yields higher NPLs in NLOS scenarios.
\begin{table}[!ht]

\centering
\caption{\footnotesize Ablation Studies on the SNR and link state terms. The metric is NPL averaged over all testing tasks.}
\label{tab:ablation}
\vspace{-1ex}
\footnotesize
\begin{tabular}{llll}
\toprule
                 & LOS                         & 1-NLOS                      & $2^+$-NLOS                     \\ \hline
WAN                 & $1.01\pm 0.01$ & $1.45\pm 0.03$  & $3.83\pm 0.81$\\ 
PiPRL                & $1.01 \pm 0.01$ & $1.41 \pm 0.03$ & $2.60 \pm 0.05$ \\ 
SNR Ablation        & $1.02 \pm 0.02$  & $1.46 \pm 0.04$  & $4.62 \pm 1.15$  \\
Link State Ablation & $1.02 \pm 0.02$ & $1.47 \pm 0.05$  & $3.90 \pm 1.02$  \\ 
\bottomrule
\end{tabular}
\end{table}

\begin{table}[!ht]
\centering
\caption{\footnotesize GPU hours and Episodes (Eps) of PiPRL and NPRL for Maps A-O. Experiments are conducted on a Linux GPU workstation with an AMD Threadripper 3990X (64 Cores, 2.90 GHz) and an NVidia RTX 8000.}
\resizebox{\linewidth}{!}{
\begin{tabular}{llccccc}
\toprule
\multirow{2}{*}{Label} & \multirow{2}{*}{Map Name} & \multicolumn{2}{c}{PiPRL} & \multicolumn{2}{c}{NPRL} & \multirow{2}{*}{\shortstack{Reduction \\ (GPU Hours\%)}} \\
\cmidrule(r){3-4} \cmidrule(r){5-6}
      &          & GPU Hours & EPs (Average) & GPU Hours & EPs (Average) & \\
\midrule
A & Bowlus      & 20.35 & 798 & 27.70 & 1000 & 26.56 \\
B & Arkansaw    & 17.81 & 579 & 25.32 & 1000 & 29.67 \\
C & Andrian     & 15.21 & 440 & 24.90 &  996 & 38.93 \\
D & Anaheim     & 13.34 & 384 & 25.11 & 1000 & 46.86 \\
E & Andover     & 5.21  & 208 & 23.50 &  940 & 77.83 \\
F & Annawan     & 5.02  & 201 & 23.32 &  933 & 78.47 \\
G & Azusa       & 4.67  & 187 & 22.11 &  884 & 78.87 \\
H & Ballou      & 4.42  & 177 & 21.27 &  851 & 79.21 \\
I & Capistrano  & 6.83  & 273 & 24.33 & 1000 & 71.93 \\
J & Delton      & 4.87  & 195 & 22.58 &  903 & 78.42 \\
K & Bolton      & 5.28  & 211 & 24.00 &  960 & 77.99 \\
L & Goffs       & 3.91  & 156 & 21.37 &  855 & 81.70 \\
M & Hainesburg  & 4.85  & 194 & 23.00 &  920 & 78.91 \\
N & Kerrtown    & 4.72  & 189 & 23.10 &  924 & 79.56 \\
O & Micanopy    & 3.93  & 157 & 23.17 &  927 & 83.03 \\
\bottomrule
\label{tab:final_gpu_episodes}
\end{tabular}
}
\vspace{-2ex}
\end{table}

\begin{table*}[!ht]
\centering
\caption{A comparison of NPLs under 6 testing maps. PIRL achieves impressively efficient navigation in the challenging scenario $2^+$-NLOS, compared with baselines. }
\label{tab:npl}
\fontsize{6.6pt}{8pt}\selectfont
\begin{tabular}{|l|ccc|ccc|ccc|}
\hline
            & \multicolumn{3}{c|}{Map P}                                                                                                                                                                                                            & \multicolumn{3}{c|}{Map Q}                                                                                                                                                                                                            & \multicolumn{3}{c|}{Map R}                                                                                                                                                                                                            \\ \hline
           & \multicolumn{1}{c|}{LOS}                                                        & \multicolumn{1}{c|}{1-NLOS} & \multicolumn{1}{c|}{$2^+$-NLOS} & \multicolumn{1}{c|}{LOS}                                                        & \multicolumn{1}{c|}{1-NLOS} & \multicolumn{1}{c|}{$2^+$-NLOS} & \multicolumn{1}{c|}{LOS}                                                        &\multicolumn{1}{c|}{1-NLOS} & \multicolumn{1}{c|}{$2^+$-NLOS} \\ \hline
PiPRL       & \multicolumn{1}{c|}{\begin{tabular}[c]{@{}c@{}}1.01 \\ $\pm$ 0.01\end{tabular}} & \multicolumn{1}{c|}{\begin{tabular}[c]{@{}c@{}}1.25 \\ $\pm$ 0.02\end{tabular}}    & \begin{tabular}[c]{@{}c@{}}2.13 \\ $\pm$ 0.04\end{tabular}     & \multicolumn{1}{c|}{\begin{tabular}[c]{@{}c@{}}1.01 \\ $\pm$ 0.01\end{tabular}} & \multicolumn{1}{c|}{\begin{tabular}[c]{@{}c@{}}1.37 \\ $\pm$ 0.02\end{tabular}}    & \begin{tabular}[c]{@{}c@{}}2.43 \\ $\pm$ 0.05\end{tabular}     & \multicolumn{1}{c|}{\begin{tabular}[c]{@{}c@{}}1.01 \\ $\pm$ 0.00\end{tabular}} & \multicolumn{1}{c|}{\begin{tabular}[c]{@{}c@{}}1.13 \\ $\pm$ 0.03\end{tabular}}    & \begin{tabular}[c]{@{}c@{}}2.65\\ $\pm$ 0.06\end{tabular}      \\ \hline
NPRL        & \multicolumn{1}{c|}{\begin{tabular}[c]{@{}c@{}}2.03 \\ $\pm$ 1.02\end{tabular}} & \multicolumn{1}{c|}{\begin{tabular}[c]{@{}c@{}}2.72 \\ $\pm$ 0.55\end{tabular}}    & \begin{tabular}[c]{@{}c@{}}4.96 \\ $\pm$ 1.37\end{tabular}     & \multicolumn{1}{c|}{\begin{tabular}[c]{@{}c@{}}2.12 \\ $\pm$ 1.00\end{tabular}} & \multicolumn{1}{c|}{\begin{tabular}[c]{@{}c@{}}3.08 \\ $\pm$ 0.68\end{tabular}}    & \begin{tabular}[c]{@{}c@{}}5.00 \\ $\pm$ 1.41\end{tabular}     & \multicolumn{1}{c|}{\begin{tabular}[c]{@{}c@{}}2.28 \\ $\pm$ 1.03\end{tabular}} & \multicolumn{1}{c|}{\begin{tabular}[c]{@{}c@{}}2.49 \\ $\pm$ 0.81\end{tabular}}    & \begin{tabular}[c]{@{}c@{}}4.99 \\ $\pm$ 1.20\end{tabular}     \\ \hline
V-SLAM    & \multicolumn{1}{c|}{\begin{tabular}[c]{@{}c@{}}1.05 \\ $\pm$ 0.02\end{tabular}} & \multicolumn{1}{c|}{\begin{tabular}[c]{@{}c@{}}1.82 \\ $\pm$ 0.51\end{tabular}}    & \begin{tabular}[c]{@{}c@{}}4.58 \\ $\pm$ 1.12\end{tabular}     & \multicolumn{1}{c|}{\begin{tabular}[c]{@{}c@{}}1.11\\ $\pm$ 0.03\end{tabular}}  & \multicolumn{1}{c|}{\begin{tabular}[c]{@{}c@{}}2.89 \\ $\pm$ 0.73\end{tabular}}    & \begin{tabular}[c]{@{}c@{}}4.89\\ $\pm$ 1.00\end{tabular}      & \multicolumn{1}{c|}{\begin{tabular}[c]{@{}c@{}}1.09 \\ $\pm$ 0.03\end{tabular}} & \multicolumn{1}{c|}{\begin{tabular}[c]{@{}c@{}}1.68\\ $\pm$ 0.6\end{tabular}}      & \begin{tabular}[c]{@{}c@{}}4.68 \\ $\pm$ 1.01\end{tabular}     \\ \hline
WAN & \multicolumn{1}{c|}{\begin{tabular}[c]{@{}c@{}}1.02 \\ $\pm$ 0.00\end{tabular}} & \multicolumn{1}{c|}{\begin{tabular}[c]{@{}c@{}}1.32 \\ $\pm$ 0.02\end{tabular}}    & \begin{tabular}[c]{@{}c@{}}3.88 \\ $\pm$ 0.82\end{tabular}     & \multicolumn{1}{c|}{\begin{tabular}[c]{@{}c@{}}1.01 \\ $\pm$ 0.01\end{tabular}} & \multicolumn{1}{c|}{\begin{tabular}[c]{@{}c@{}}1.63 \\ $\pm$ 0.05\end{tabular}}    & \begin{tabular}[c]{@{}c@{}}3.71 \\ $\pm$ 0.71\end{tabular}     & \multicolumn{1}{c|}{\begin{tabular}[c]{@{}c@{}}1.01 \\ $\pm$ 0.00\end{tabular}} & \multicolumn{1}{c|}{\begin{tabular}[c]{@{}c@{}}1.23 \\ $\pm$ 0.02\end{tabular}}    & \begin{tabular}[c]{@{}c@{}}3.97 \\ $\pm$ 0.83
\end{tabular}   \\ \hline
           & \multicolumn{3}{c|}{Map S}                                                                                                                                                                                                            & \multicolumn{3}{c|}{Map T}                                                                                                                                                                                                            & \multicolumn{3}{c|}{Map U}                                                                                                                                                                                                            \\ \hline
           & \multicolumn{1}{c|}{LOS}                                                        & \multicolumn{1}{c|}{1-NLOS} & \multicolumn{1}{c|}{$2^+$-NLOS} & \multicolumn{1}{c|}{LOS}                                                        & \multicolumn{1}{c|}{1-NLOS} & \multicolumn{1}{c|}{$2^+$-NLOS} & \multicolumn{1}{c|}{LOS}                                                        & \multicolumn{1}{c|}{1-NLOS} & \multicolumn{1}{c|}{$2^+$-NLOS} \\ \hline
PIRL       & \multicolumn{1}{c|}{\begin{tabular}[c]{@{}c@{}}1.01 \\ $\pm$ 0.01\end{tabular}} & \multicolumn{1}{c|}{\begin{tabular}[c]{@{}c@{}}1.23 \\ $\pm$ 0.01\end{tabular}}    & \begin{tabular}[c]{@{}c@{}}2.82 \\ $\pm$ 0.04\end{tabular}     & \multicolumn{1}{c|}{\begin{tabular}[c]{@{}c@{}}1.00\\ $\pm$ 0.00\end{tabular}}  & \multicolumn{1}{c|}{\begin{tabular}[c]{@{}c@{}}1.43 \\ $\pm$ 0.03\end{tabular}}    & \begin{tabular}[c]{@{}c@{}}2.59\\ $\pm$ 0.06\end{tabular}      & \multicolumn{1}{c|}{\begin{tabular}[c]{@{}c@{}}1.01 \\ $\pm$ 0.01\end{tabular}} & \multicolumn{1}{c|}{\begin{tabular}[c]{@{}c@{}}1.73 \\ $\pm$ 0.03\end{tabular}}    & \begin{tabular}[c]{@{}c@{}}2.46\\ $\pm$ 0.05\end{tabular}      \\ \hline
NPRL        & \multicolumn{1}{c|}{\begin{tabular}[c]{@{}c@{}}2.01 \\ $\pm$ 0.99\end{tabular}} & \multicolumn{1}{c|}{\begin{tabular}[c]{@{}c@{}}2.81 \\ $\pm$ 0.83\end{tabular}}    & \begin{tabular}[c]{@{}c@{}}5.14 \\ $\pm$ 1.21\end{tabular}     & \multicolumn{1}{c|}{\begin{tabular}[c]{@{}c@{}}2.23\\ $\pm$ 1.10\end{tabular}}  & \multicolumn{1}{c|}{\begin{tabular}[c]{@{}c@{}}3.13 \\ $\pm$ 0.83\end{tabular}}    & \begin{tabular}[c]{@{}c@{}}4.88 \\ $\pm$ 1.86\end{tabular}     & \multicolumn{1}{c|}{\begin{tabular}[c]{@{}c@{}}1.90\\ $\pm$ 0.89\end{tabular}}  & \multicolumn{1}{c|}{\begin{tabular}[c]{@{}c@{}}3.25 \\ $\pm$ 0.64\end{tabular}}    & \begin{tabular}[c]{@{}c@{}}4.50\\ $\pm$ 1.05\end{tabular}      \\ \hline
V-SLAM    & \multicolumn{1}{c|}{\begin{tabular}[c]{@{}c@{}}1.04\\ $\pm$ 0.03\end{tabular}}  & \multicolumn{1}{c|}{\begin{tabular}[c]{@{}c@{}}1.99 \\ $\pm$ 0.58\end{tabular}}    & \begin{tabular}[c]{@{}c@{}}4.98 \\ $\pm$ 1.00\end{tabular}     & \multicolumn{1}{c|}{\begin{tabular}[c]{@{}c@{}}1.10 \\ $\pm$ 0.08\end{tabular}} & \multicolumn{1}{c|}{\begin{tabular}[c]{@{}c@{}}3.01 \\ $\pm$ 0.67\end{tabular}}    & \begin{tabular}[c]{@{}c@{}}4.69 \\ $\pm$ 1.01\end{tabular}     & \multicolumn{1}{c|}{\begin{tabular}[c]{@{}c@{}}1.06\\ $\pm$ 0.04\end{tabular}}  & \multicolumn{1}{c|}{\begin{tabular}[c]{@{}c@{}}3.19\\ $\pm$ 0.56\end{tabular}}     & \begin{tabular}[c]{@{}c@{}}4.43\\ $\pm$ 1.00\end{tabular}      \\ \hline
WAN & \multicolumn{1}{c|}{\begin{tabular}[c]{@{}c@{}}1.01 \\ $\pm$ 0.00\end{tabular}} & \multicolumn{1}{c|}{\begin{tabular}[c]{@{}c@{}}1.32 \\ $\pm$ 0.03\end{tabular}}    & \begin{tabular}[c]{@{}c@{}}3.78 \\ $\pm$ 0.90\end{tabular}     & \multicolumn{1}{c|}{\begin{tabular}[c]{@{}c@{}}1.00 \\ $\pm$ 0.00\end{tabular}} & \multicolumn{1}{c|}{\begin{tabular}[c]{@{}c@{}}1.48 \\ $\pm$ 0.02\end{tabular}}    & \begin{tabular}[c]{@{}c@{}}4.01 \\ $\pm$ 0.90\end{tabular}     & \multicolumn{1}{c|}{\begin{tabular}[c]{@{}c@{}}1.01 \\ $\pm$ 0.01\end{tabular}} & \multicolumn{1}{c|}{\begin{tabular}[c]{@{}c@{}}1.74 \\ $\pm$ 0.04\end{tabular}}    & \begin{tabular}[c]{@{}c@{}}3.63 \\ $\pm$ 0.70\end{tabular}     \\ \hline
\end{tabular}
\end{table*}

\newpage
\bibliography{main}
\bibliographystyle{rlj}


\end{document}